\pdfoutput=1
\documentclass[11pt]{article}

\usepackage[]{acl}

\usepackage{times}
\usepackage{latexsym}

\usepackage[T1]{fontenc}
\usepackage[utf8]{inputenc}

\usepackage{microtype}

\usepackage{inconsolata}

\usepackage{cleveref}
\usepackage{multirow}
\usepackage{arydshln}
\usepackage{graphicx}
\graphicspath{ {./figures/} }
\usepackage{booktabs}
\usepackage{pgfplots,tikz}
\pgfplotsset{compat=1.16}
\usetikzlibrary{intersections, positioning, calc, math}
\usepackage{xcolor}
\definecolor{azure}{rgb}{0.0, 0.5, 1.0}

\title{Memory Augmented Language Models through \\ Mixture of Word Experts}

\author{ Cicero Nogueira dos Santos, James Lee-Thorp, Isaac Noble \\
  {\bf Chung-Ching Chang, David Uthus } \\
  Google Research \\
  \texttt{cicerons,jamesleethorp,isaacn,ccchang,duthus@google.com}\\
}

\begin{document}
\maketitle
\begin{abstract}
Scaling up the number of parameters of language models has proven to be an effective approach to improve performance. For dense models, increasing model size proportionally increases the model's computation footprint. In this work, we seek to aggressively decouple learning capacity and FLOPs through Mixture-of-Experts (MoE) style models with large knowledge-rich vocabulary based routing functions and experts.
Our proposed approach, dubbed Mixture of Word Experts (MoWE), can be seen as a memory augmented model,
where a large set of word-specific experts play the role of a sparse memory. 
We demonstrate that MoWE performs significantly better than the T5 family of models with similar number of FLOPs in a variety of NLP tasks.
Additionally, MoWE outperforms regular MoE models on knowledge intensive tasks and has similar performance to more complex memory augmented approaches that often require to invoke custom mechanisms to search the sparse memory.
\end{abstract}

\section{Introduction}

Increasing the parameter count of language models has been a primary driver of increased model quality \cite{2020t5,kaplan2020scaling,brown2020language}.
This is particularly apparent on knowledge intensive tasks, such as TriviaQA \cite{joshi-etal-2017-triviaqa}, where language models with more parameters and learning capacity benefit from soaking up world knowledge from their pretraining data \cite{chowdhery2022palm,touvron2023llama}. However, increasing the model size also increases the cost of running the model.

\begin{figure}[ht!]
    \centering
    \footnotesize
    \begin{tikzpicture}
    \begin{axis}[
        xlabel={\# Float Operations per Target Token},
        ylabel={TriviaQA Exact Match},
        xmin=180000000,
        xmax=40000000000,
        ymin=20, ymax=50.0,
        ytick distance=5,
        xmode=log,
        xtick={500000000, 1000000000, 5000000000, 10000000000},
        xticklabels={$5 \cdot 10^{8}$, $10^{9}$, $5 \cdot 10^{9}$, $10^{10}$},
        x tick label style={rotate=30,anchor=north east,},
        ymajorgrids=true,
        grid style=dashed,
        axis lines = left,
        width=\linewidth,
        height=4.5cm,
        every node near coord/.append style={yshift=3pt},
        scatter, mark=*,
        scatter src=explicit symbolic,
        scatter/classes={a={azure}, b={red}},
    ]
    
    \newcommand\modelsmall{\fontsize{8pt}{11pt}\selectfont}

    \addplot[color=azure, scatter, mark size=1.5pt, style={font=\modelsmall},
        nodes near coords*={\Label},
        visualization depends on={value \thisrow{label} \as \Label},
    ] table [meta=class] {
        x y class label
        396791484.6 39.4 a MoWE-Base
        1409880280 44.8 a MoWE-Large
        };
        
    \addplot[color=red, scatter, mark size=1.5pt, dash pattern=on 2pt off 2pt,mark=square*, style={font=\modelsmall},
            nodes near coords*={\Label},
        visualization depends on={value \thisrow{label} \as \Label},
    ]
        table [meta=class] {
        x y class label
        396791484.6 24.2 b T5-Base
        1409880280 28.5 b T5-Large
        5219343953 36.0 b T5-XL
        20589786597 42.9 b T5-XXL
        };

    \end{axis}
    \end{tikzpicture}    
    
    \caption{\textbf{MoWE vs T5.1.1 on TriviaQA}: MoWE-Base and MoWE-Large perform as well as T5.1.1-XL and T5.1.1-XXL, respectively, while using a significantly smaller number of FLOPs. T5.1.1 results are from \cite{roberts-etal-2020-much}.}
    \label{fig:mowe_vs_t5_flops}
\end{figure}
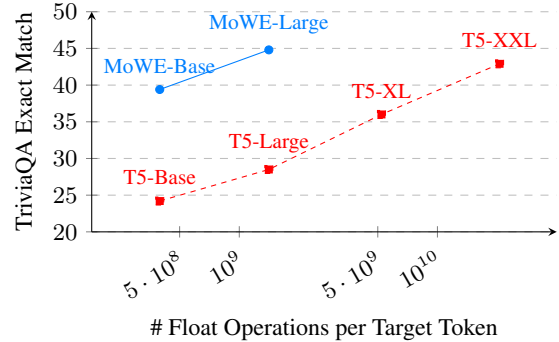

\begin{figure*}[ht!] 
    \centering
    \includegraphics[scale=.5]{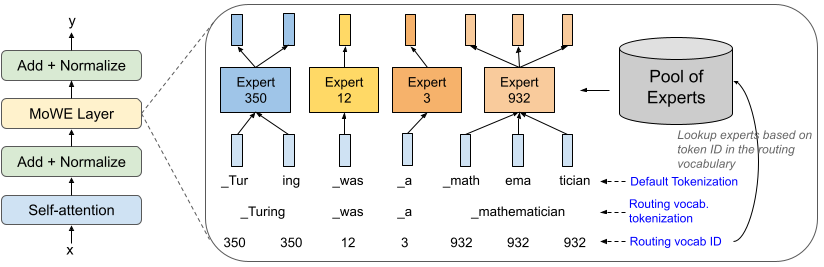}
    \caption{\textbf{MoWE Layer}: We replace the FFN layer in a subset of Transformer blocks by a \textit{MoWE Layer}, which is a sparse layer that  processes tokens using multiple experts (FFNs). Each input token is processed by a single expert that is selected based on the input token id (at the corresponding sequence position) in the routing vocabulary.}
    \label{fig:mowe_block}
\end{figure*}

In this work, we build on the Mixture-of-Experts (MoE) paradigm to design a neural net architecture that enjoys the quality benefits from scaling the parameter count but remains FLOPs and latency efficient.
Our proposed approach, which we name Mixture-of-Word-Experts (MoWE), follows two design principles:
(1) a very large number of experts (tens of thousands instead of 32 to 128 normally used in MoEs) that
(2) are "word-specific" -- that is, they are tied to a large knowledge-rich vocabulary through fixed routing function.
The core MoWE layer is illustrated in Figure \ref{fig:mowe_block}.
MoWE models are memory augmented models, where the large set of word experts (small MLPs) play the role of a sparse memory that is seamlessly integrated to the main model backbone.

We empirically demonstrate that MoWE significantly outperforms T5 models \cite{2020t5} with a comparable number of FLOPs across a variety of NLP tasks.
Focusing on knowledge intensive tasks such as TriviaQA \cite{joshi-etal-2017-triviaqa} and WebQuestions \cite{berant-etal-2013-webquestions}, we show that a
MoWE "Base" sized outperforms T5-XL and a MoWE "Large" outperforms T5-XXL models (see Figure \ref{fig:mowe_vs_t5_flops}),
while being at least 4.3x and 6.6x faster to train, respectively.
MoWE outperforms vanilla MoE models \cite{shazeer2017,lepikhin2020gshard,fedus2022switch} on knowledge intensive tasks, while matching performance on NLP task suites such as SuperGLUE \cite{wang2019superglue}.
Additionally, MoWE also matches or outperforms recently proposed knowledge augmented models \cite{fevry-etal-2020-entities,jong2022mention}, while avoiding invoking any custom mechanism to search the sparse memory.

In summary, the main contributions of this work are:
\begin{itemize}
    \item We propose a novel neural net architecture that effectively combines the efficiency of sparse models with the power of large language models to memorize and retrieve world knowledge; see Table \ref{tab:mowe_memory} for a downstream peak at how these memories are used.
    \item We introduce very large auxiliary vocabularies to perform routing.
    \item We propose and validate a new strategy to efficiently train MoE models with: (1) hundreds of thousands of experts and (2) very unbalanced token assignments across experts. 
    \item For knowledge intensive tasks such as question answering and claim verification, we present new efficient sparse models that outperform larger, significantly slower dense models that use an order of magnitude more FLOPs.
\end{itemize}

\section{Mixture-of-Word-Experts}

\subsection{Mixture-of-Experts (MoE) Background}
Transformer-based MoE architectures \cite{lepikhin2020gshard,du2022glam,fedus2022switch} are implemented by replacing the dense Feed Forward Network (FFN) layer in a subset of Transformer blocks with a sparse layer of experts.
Instead of using a single FFN to process all inputs,
the sparse layer employs a set of FFNs (the experts). Each token representation is processed by a single (top-1) or a subset (top-k) of experts.
The promise in MoE models is to vastly increase the number of parameters in the network without significantly increasing the amount of computation.

Common MoE implementations replace every other FFN layer of the Transformer architecture by a sparse layer that contains between 32 and 128 experts \cite{lepikhin2020gshard,du2022glam,fedus2022switch}.
Tokens are assigned to particular experts by a \textit{routing function} that is learned jointly with the rest of the parameters of the network. Because of the nature of the one-hot assignments of tokens to experts, training the routing function is tricky and typically performed indirectly by rescaling expert outputs by the assignment probability (the "router confidence") that a given token should be assigned to a particular expert. 

% An auxiliary loss \cite{shazeer2017} is used to encourage the router to evenly distribute tokens across experts.
% Additionally, to avoid instabilities during training, additional tricks such as the \textit{router z-loss} \cite{zoph2022stmoe} are employed.
%Learning the routing function is not easy, and sometimes it does not align well between the pretraining and downstream applications.
%ometime it is easy to get very good gains in the upstream (pretraining) task while not getting gains in the downstream tasks.

\subsection{Mixture-of-Word-Experts (MoWE) Architecture}
Similar to MoE models, MoWE is a Transformer-based architecture \cite{transformer} where the FFN layer of a subset of Transformer blocks is replaced by a \textit{MoWE Layer},
which is a sparse layer that processes tokens using a pool of experts (FFNs).
In a MoWE layer, a token representation at position \emph{i} is processed by a single expert that is selected based on the id, in the \textit{routing vocabulary}, of the corresponding \emph{input} sequence token at position \emph{i}. Figure \ref{fig:mowe_block} illustrates a MoWE layer.

\textbf{Routing decisions are driven by a large auxiliary vocabulary}.
There are two tokenizations of the input:
(1) the \textit{default tokenization} which is the regular one that defines the input tokens and their embeddings; and 
(2) the \textit{routing tokenization},
which is performed using a large auxiliary \textit{routing vocabulary} (introduced in Section \ref{sec:vocab}). 
The token ids resulting from the routing tokenization are called \textit{routing ids}.
In a MoWE layer,
routing consists of mapping routing ids to experts ids through a hash function. 
In the extreme case where each word in the routing vocabulary has its own expert, the routing id corresponds directly to the expert id, as illustrated in Figure \ref{fig:mowe_block}.

\textbf{Importance of a large pool of experts}.
A MoWE layer uses tens or hundreds of thousands of experts, which are normally smaller (smaller MLP dimension) than the regular, dense FFN layer.
The goal of using a large number of experts is to encourage specialization.
With an extremely large number of experts,
each word in the routing vocabulary is assigned to its own expert.
However, we found that it is more efficient (both in terms of memory and training signal) to have fewer experts than vocabulary entries and share some experts across multiple routing ids.
Nevertheless,
a token with a given id is always routed to the same expert.

Recent work suggests that Transformers act as key-value memories \cite{geva-etal-2021-ffn_memory,dai-etal-2022-knowledge_neurons,zhang-etal-2022-moefication}, 
and that factual knowledge seems to be stored in the FFNs \cite{dai-etal-2022-knowledge_neurons,meng2022locating}.
We conjecture that the large routing vocabulary and associated large number of experts further encourage the MoWE layer to function as a sparse memory.
We find that using complete words instead of word pieces (see Section \ref{sec:vocab}) to perform routing is a strong inductive bias that makes it easier for the experts to specialize on specific words.
For example, the expert for the word ``{Turing}'' will be activated only when that word appears in the input, and therefore will be specialized on content that co-occur with that word.
By using word-specific \emph{key-value memories} (word experts), our hope is that MoWE can make it easier for the model to store and retrieve information about those words.

% Using a large number of experts (in some cases one per word), we can enforce the expert to store more information about the context where that specific word appears.
% When the number of experts is smaller than the routing vocabulary, the same expert will process tokens of different ids.

\subsection{Overcoming the Challenges of using Tens of Thousands of Experts}

\begin{figure}[h!]
    \centering
    \includegraphics[scale=.46]{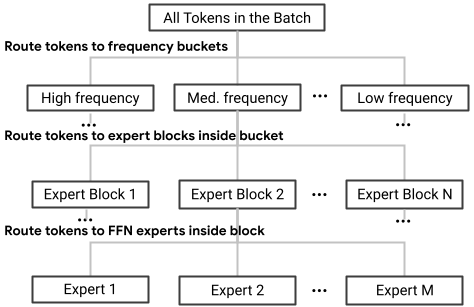}
    \caption{\textbf{Hierarchical Routing.} Tokens are first routed to buckets that handle routing ids of similar frequency.
    Inside each bucket, experts are grouped in blocks, and each token is routed to the block that contains its assigned expert.
    Inside the block, each token is routed to and processed by an actual expert.}
    \label{fig:hierarchical_routing}
\end{figure}

Most large scale MoE models are implemented using the single program, multiple data (SPMD) parallelism strategy; see, for example, \cite{lepikhin2020gshard}.
Data and experts are cosharded across devices. Data that is originally on device \emph{x} but is assigned, by the routing function, to an expert on device \emph{y} must be transferred between devices through all-to-all communications.
Under the single program paradigm on modern accelerators, experts send and receive the same amount of data and perform that same amount of computation (same array shapes on each device). Effectively implementing MoWE using vanilla SPMD poses some key challenges: 
(1) The sheer number of experts brings an unpractical overhead in terms of all-to-all communication.
(2) Word frequency follows a Zipfian-like distribution. This unbalanced nature of vocabulary-driven routing requires different word experts to process orders of magnitude more tokens than others. 
% And this only gets worse when using approximately one expert per word.
We propose a new strategy that overcomes these challenges and allows an efficient implementation of the MoWE layer.
Our method contains three main ingredients:

\textbf{Expert Blocks.} We group experts into blocks that are sharded across devices. All-to-all communication is only performed between blocks instead of between experts.
Provided we keep the number of expert blocks small enough, we can increase the number of experts without increasing all-to-all communication costs.
For example, if we use 128 blocks with 256 experts each, we end up with 32768 experts.
We are able to use expert blocks because the fixed routing function pre-defines which block, and which expert inside the block, will process a given token.

\textbf{Frequency Bucketing.}
To overcome the unbalanced word frequency distribution, we compute the frequency of words in a sample of 2B tokens from our pretraining data and then split the routing vocabulary into $k$ buckets,
where the words in each bucket have approximately the same frequency.
Each bucket is then handled by a separate set of expert blocks.
% The computational graph is split into $k$ branches that work like separate SPMD programs.
Conceptually, the $k$ MoWE layers are executed in parallel.
With this approach, experts in different buckets can have different sizes or even different architectures and can support different token capacities (process a different number of tokens)\footnote{\citet{gale2023megablocks} offers a potential way to avoid some of this bucket, although there likely remains similar effective lower bounds to array and bucket shapes to ensure efficiency.}.

\textbf{Hierarchical Routing.} Given a batch of tokens, the first step is to route them to frequency buckets.
Next, inside each bucket, each token is routed to the expert block that contains its assigned expert.
Finally, inside the block, each token is routed to and processed by an actual expert.
Since routing decisions are based purely on (static) routing ids,
token-to-expert assignments are known beforehand and the full path through the hierarchical routing tree becomes trivial.
Fig. \ref{fig:hierarchical_routing} illustrates this process.

Our proposed strategy allowed us to pretrain MoWE-Base models with up to 1 million (small) experts using 16 v3 TPUs.
We did not observe any training instability (e.g. gradient blowup) that are often reported in the pretraining of regular MoE models \cite{zoph2022stmoe}; we suspect is a helpful artifact of our fixed routing scheme.

\subsection{Knowledge-Rich Routing Vocabulary}
\label{sec:vocab}
A straightforward strategy to build a large routing vocabulary consists in using the pretraining dataset to train a large vocabulary SentencePiece tokenizer \cite{kudo-richardson-2018-sentencepiece}.
However,
initial experiments indicated that this method is suboptimal as many words in the vocabulary turn out to be uninformative --  many are just variations of the form of other words.
To build a \emph{knowledge-rich routing vocabulary} that contains more \emph{informative} tokens, we derive the vocabulary from a knowledge rich dataset as follows:
\begin{itemize}
    \item[(1)] Start with the set of all entity and relation names that appears in a Wikidata dump.\footnote{The routing vocabulary used in our experiments was derived from the Wikidata dump previously used by  \citet{agarwal-etal-2021-knowledge}.}
    \item[(2)] Lowercase and split each name using white space and a regex to remove punctuation.\footnote{Languages that do not use white space for word splitting will require slightly modified processing.}
    \item[(3)] Order tokens based on their frequency in the C4 dataset \cite{2020t5} (version 2.2.0), which is our pretraining dataset.
    \item[(4)] Select the top 1M tokens to form our routing vocabulary.
\end{itemize}
This strategy increases the likelihood that the majority of entries in the vocabulary are (single word) names -- i.e., terms that we want to store knowledge about. For example, tokenization with a T5.1.1 32K vocabulary breaks down the word ``\texttt{mathematician}'' into 5 tokens (``\texttt{math}'', ``\texttt{e}'',``\texttt{m}'',``\texttt{a}'', ``\texttt{tician}''),
while our 1M routing vocabulary keeps it as a single token; see also Figure \ref{fig:mowe_block}.
Ideally, the two tokenizations should be aligned as in the figure, but the only hard constraint is that each token from the default tokenization (which defines the input sequence) needs to have a routing id.
Appendix \ref{sec:appendix:knowledge_vocab} shows more samples of the top words in the routing vocabulary. 

Finally,
to allow 
(a) efficient lookup of routing ids
 and (b) the use of the MoWE layer in auto-regressive scenarios where normally only the initial part of the word is known,
we approximate the \textit{routing tokenization} using a hash operation.
More specifically, we use the following steps:
\begin{itemize}
    \item \textbf{Offline:} (1) we extend the auxiliary vocabulary by concatenating the default T5 32K vocabulary to it. (2) we tokenize each entry in the auxiliary vocabulary using the default tokenizer and build a hash table where the key is the sequence of (default) token ids and the value is the routing id (a sequential number).
    \item \textbf{Online:} given a tokenized input sequence $s$ composed of $n$ token ids $\{t_1, t_2, ...,t_n\}$, we create the routing id of token $t_i$ by first looking up in the hash-table all sub-sequences $\{t_{i-k}, ...,t_i\}$ for $k \in [0, 8]$, and adopt the routing id of the largest sub-sequence.
\end{itemize}

\section{Experimental Setup}

\begin{table*}[h!]
\small
\centering
\begin{tabular}{lcccccc}
\toprule
\textbf{Model} & \textbf{TriviaQA} & \textbf{WebQuestions} & \textbf{Natural Questions} & \textbf{FEVER} & \textbf{SuperGLUE} & \textbf{Train Time Ratio} \\
& & & & & & \textbf{to T5.1.1-Base} \\
\midrule
T5.1.1-Base  & 24.2 & 28.2 & 25.7 & 61.3 & 77.2 & 1.0\\
MoWE-Base & 39.4 & 35.7 & 29.6 & 66.3 & 83.5 & 2.0 \\
\midrule
T5.1.1-Large & 28.2 & 29.5 & 27.3 & 63.0 & 85.1 & 3.1 \\
MoWE-Large & \textbf{44.8} & \textbf{38.8} & 31.9 & \textbf{68.5} & 87.4 & 4.0 \\
\midrule
T5.1.1-XL & 36.0 & 32.4 & 29.5 & 65.9 & 88.5 & 8.6 \\
T5.1.1-XXL & 42.9 & 35.6 & \textbf{32.8} & 67.5 & \textbf{89.9} & 26.4 \\
\bottomrule
\end{tabular}
\caption{Comparison of MoWE and T5.1.1 models on five different language processing tasks. We use exact match for TriviaQA, WebQuestions and Natural Questions. We use accuracy for FEVER and a blended average of accuracy and F1 scores for the SuperGLUE suite as in \cite{2020t5}.
T5.1.1. results for TriviaQA, WebQuestions and Natural Questions are from \cite{roberts-etal-2020-much}. For each model, we also report the training time relative to T5.1.1-Base.; estimated by running each model with a batch size of 256 and input (output) sequence length of 512 (62) on 64 v3 TPUs -- the smallest slice that could fit T5-XXL with 256 examples. Note that this likely underestimates the speed of the smaller models, which would enjoy better utilization on fewer devices.}
\label{tab:results:mowe_vs_t5}
\end{table*}

\subsection{Tasks and Datasets}
We present results on a wide range of NLP tasks.
That said, as our main goal is to assess the performance of MoWE on knowledge intensive tasks, we focus our analysis on closed-book question answering tasks: 
TriviaQA \cite{joshi-etal-2017-triviaqa},
WebQuestions \cite{berant-etal-2013-webquestions}
and Natural Questions \cite{kwiatkowski-etal-2019-natural}.
As in \citet{roberts-etal-2020-much}, our model has no access to external knowledge/text during finetuning and inference.
Similar to \citet{lee-etal-2019-latent,roberts-etal-2020-much}, we perform evaluation by holding out 10\% of the training set as a validation set; models are finetuned on the remaining 90\% of the data.

We also check the performance of MoWE for the claim verification task using the FEVER dataset \cite{thorne-etal-2018-fever},
which contains separate validation and test sets.
Finally, 
to compare our results with classic MoE Transformer models \cite{lepikhin2020gshard}, we apply MoWE to SuperGLUE benchmark \cite{superglue}. We pretrain all models on the C4 dataset \cite{2020t5}, version 2.2.0.

\subsection{MoWE Setup and Hyperparameters.}
Following popular \cite{fedus2022switch} and state-of-the-art \cite{zoph2022stmoe} Transformer-based encoder-decoder MoE models, we use T5.1.1 as the backbone of our MoWE models.

Our main results are from an architecture with four MoWE-layers --
two in the encoder and two in the decoder, and each MoWE layer contains 32K experts.
We use four MoWE layers as they offer good accuracy without sacrificing computational performance due to routing overhead (see Appendix \ref{tab:results:num_mowe_layers}).
We place MoWE layers near the middle of the encoder (decoder) to ensure that: 
(1) the MoWE layers receive a representation of the token that is already somewhat contextualized;
(2) after the MoWE layer, there are still multiple Transformer Blocks that can benefit from the output of that layer.
Parameters are shared across all MoWE layers with the following goal:
(1) it makes the MoWE layer even more similar to a memory that is accessed at different points of the network;
(2) we can keep the overall number of sparse parameters relatively low without the need to decrease the total and the size of experts.
Additionally, empirical results indicated that sharing parameters across the MoWE layers leads to better performance.
The routing vocabulary has 2$^{20}$ ($\sim$1M) entries and was constructed as described in Section \ref{sec:vocab}.
MoWE-Base and MoWE-Large models have 31B and 45.5B parameters, respectively.
See Appendix \ref{sec:appendix:mowe_setup} for more details.

Pretraining is performed using the same span masking approach used in T5 \cite{2020t5}.
Following T5 models, our main results use MoWE models pretrained for roughly 1 trillion tokens -- 1M steps, with batch size 2048 and input sequence length of 512 tokens; the target sequence length is 114.
We use the  same pretraining hyperparameters of T5.1.1, and use 64 TPUs v3 for pretraining.

\begin{table*}[h!]
\small
\centering
\begin{tabular}{lccccccccc}
\toprule
\textbf{Model} & \textbf{\# of sparse } & \textbf{\# of experts} & \textbf{Avg. expert} & \textbf{\# of params} & \textbf{Params} &  \textbf{TQA} & \textbf{WQ} & \textbf{NQ} & \textbf{SG} \\
& \textbf{layers} & \textbf{per layer} & \textbf{MLP dim.} & & \textbf{shared?} \\
\midrule
MoE-Top2 & 12 & 32 & 2048 & 2B & No & 26.5 & 27.7 & \textbf{25.8} & 80.2 \\
MoWE & 4 & 8K & 141  & 2B & Yes &  \textbf{29.8} & \textbf{31.6} & \textbf{26.0} & \textbf{81.2} \\
\midrule
MoE-Top2  & 12  & 512 & 2048  & 29.2B & No & 36.2 & 31.6 & 28.5 & \textbf{83.5} \\
MoWE & 4 & 32K & 577  & 31B & Yes & \textbf{39.4} & \textbf{35.7} & \textbf{29.6} & \textbf{83.5} \\
\bottomrule
\end{tabular}
\caption{Comparison of MoWE-Base with regular MoE models on TriviaQA (TQA), WebQuestions (WQ), Natural Questions (NQ) and SuperGLUE (SG). MoE-Top2 models are based on the canonical GShard Top-2 MoE Transformer \cite{lepikhin2020gshard}.}
\label{tab:results:mowe_vs_moe}
\end{table*}

During finetuning for downstream tasks, we freeze all MoWE experts to avoid both overfitting and catastrophic forgetting of knowledge acquired during pretraining (See Appendix \ref{sec:appendix:freezing} for ablations).
This is an important distinction to MoE models, which finetune the experts for the downstream tasks.
The main hyperparameter that we tune during finetuning is the learning rate. We only use cross-entropy loss; no additional auxiliary losses are used.
\section{Experimental Results and Discussion}

\subsection{Comparison with T5.1.1}
In Table \ref{tab:results:mowe_vs_t5},
we summarize MoWE results on 5 different NLP tasks and alongside T5.1.1 models. MoWE-Base and MoWE-Large outperform T5.1.1-Base and T5.1.1-Large, respectively, on all five tasks.
There is a significant gain in performance for knowledge intensive tasks -- in particular for TriviaQA, WebQuestions and FEVER.
On TriviaQA, MoWE-Base outperforms T5.1.1-Base by 15.2 points in exact match, which corresponds to a 62.8\% improvement.
On the same dataset, MoWE-Large outperforms T5.1.1-Large by about 16.6 points.

Remarkably, MoWE-Base outperforms T5.1.1-XL on all  knowledge intensive tasks, while achieving a 4.3x relative training speedup.
Similarly, on the same tasks, MoWE-Large outperforms or has competitive results to T5.1.1-XXL, while achieving a 6.6x relative training speedup.

\subsection{Comparison with Regular MoEs}
\label{sec:comparison_to_moe}

Table \ref{tab:results:mowe_vs_moe} compares MoWE models with the canonical GShard Top-2 MoE Transformer \cite{lepikhin2020gshard}.
We use T5-Base as the backbone for all models in Table \ref{tab:results:mowe_vs_moe},
hence they have \# FLOPS similar to T5-Base.
All models in the table are trained for 1M steps with batch size 2048.
Table \ref{tab:results:mowe_vs_moe} also highlights some architectural differences between MoWE and regular MoEs. Regular MoEs use a larger number of sparse layers, each with a small number of experts and there is no parameter sharing across layers.
In MoWE, as experts are tied to the routing vocabulary and we want to encourage expert specialization,
we use a large number of experts.
Sharing expert parameters across the MoWE layers allows the use of a large number of experts without exploding the total number of parameters.

In the top part of Table \ref{tab:results:mowe_vs_moe},
we compare MoWE with a typical MoE-Top2 architecture where every other layer is sparse and each sparse layer contains 32 experts, resulting in a model of 2B parameters; see Appendix \ref{sec:appendix:metric_baselines} for details.
In order to fairly compare MoWE with this model,
we created a version of MoWE-Base that contains 2B parameters by reducing the number of experts from 32K to 8K and decreasing the expert size; see Appendix \ref{appendix:freq_bucket} for details.
At 2B scale, MoWE outperforms MoE-Top2 for all four tasks. 
In the bottom part of Table \ref{tab:results:mowe_vs_moe},
we compare our 31B sized MoWE-Base model with a version of MoE-Top2 that uses 512 experts per sparse layer and contains 29.2B params.
MoWE performs significantly better on the knowledge intensive tasks, while achieving similar performance on SuperGLUE.
We believe the superior performance of MoWE for knowledge intensive tasks comes from our strategy of using large knowledge-rich vocabulary to perform routing, as further explored in the ablations presented in Sec \ref{sec:effect_knowledge_vocab}.

\subsection{The MoWE Layer is a Sparse Memory}

We perform an experiment to assess to what extent a MoWE model relies on the MoWE layer to perform the TriviaQA task.
In particular,
we are interested in measuring the impact of deactivating the experts of relevant words when the model is generating the answer.
We then finetune this model in one of two modes:
(1) \emph{all experts activated}:
this is our regular finetuning and inference setup where all input tokens are processed by their respective experts in the MoWE layer;
(2) \emph{some experts deactivated}:
 we deactivate the experts of tokens with routing ids $>$32K during finetuning and inference.
 We set the threshold to 32K because the first 32K routing ids roughly correspond to frequent and less knowledge-driven tokens that resulted from concatenating the default vocabulary to the auxiliary one (see Section \ref{sec:vocab} for more details). 

\begin{table}[h!]
\small
\centering
\begin{tabular}{lc}
\toprule
\textbf{Selectively} & \textbf{TriviaQA EM} \\
\textbf{Deactivate Experts} & \\
\midrule
No  & \textbf{35.1} \\  
Yes & 25.6 \\ 
\bottomrule
\end{tabular}
\caption{Effect on TriviaQA exact match of deactivating experts of tokens with routing id $>$ 32K.}
\label{tab:results:deactivate_experts}
\end{table}

\begin{table*}[h!]
\small
\centering
\begin{tabular}{lcc}
\toprule
\textbf{Question} & \multicolumn{2}{c}{\textbf{Deactivate experts of highlighted}} \\ 
 &  \multicolumn{2}{c}{\textbf{words when generating answer?}} \\ 
 & \textbf{No} & \textbf{Yes} \\
\midrule
What is the name of {\color{blue} \textbf{Adele}}'s first album?   & 19 & Addiction \\
Who followed William {\color{blue} \textbf{Taft}} as US President? & Woodrow Wilson & James Garfield \\
{\color{blue} \textbf{Quinsy}} affects which part of the human body? & Tonsils & Feet \\
What country will host the {\color{blue} \textbf{2022}} FIFA World Cup competition? & Qatar & Brazil \\
What is {\color{blue} \textbf{Neptune}}'s main satellite? & Triton & Uranus \\
What was the first name of Italian {\color{blue} \textbf{statesman}} and writer {\color{blue} \textbf{Machiavelli}}? & Niccolo & Francois \\
{\color{blue} \textbf{Almeria}}, {\color{blue} \textbf{Merlot}}, and {\color{blue} \textbf{Waltham}} Cross are which fruit? & Grapes & Apple\\
\bottomrule
\end{tabular}
\caption{Example TriviaQA questions and their respective answers from two configurations of a pretrained MoWE-Base model depending on whether we deactivate the expert corresponding to routing id of highlighted words.
The answer generated by the model can change completely (from correct to incorrect in these cases) by simply deactivating the MoWE expert of a single relevant word. In this experiment, the MoWE model has a single MoWE-layer that is located in the encoder and contains 32K experts.}
\label{tab:mowe_memory}
\end{table*}

Table \ref{tab:results:deactivate_experts} shows the performance of MoWE for setups (1) and (2). 
There is a significant drop of 9 points in EM when experts of words with routing id $>$32K are deactivated.
This result indicates that MoWE models rely heavily on the experts of words that are in our knowledge-rich vocabulary.
In Table \ref{tab:mowe_memory},
we show some selected examples of questions and their respective answers for the two setups.
Deactivating a single expert\footnote{We are deactivating lots of experts in the model for setup (2), but only a single expert is used in setup (1) for each of these examples.} makes the model answer the question in a completely different way. 
For the MoWE model used in this experiment, a single expert represents only 0.33\% of the estimated total number of activated parameters.
Note that, because the MoWE layer is frozen during finetuning, 
all the knowledge that is being leveraged in the downstream task comes from the pretraining corpus.
These results suggest that (at least part of) the pretraining world knowledge needed to answer some questions is stored in the deactivated experts.
% Overall, this experiment indicates that MoWE indeed works as sparse memory.

\subsection{Comparison with Memory Augmented models}
In this section we compare the performance of MoWE with recently proposed memory augmented models: Entities as Experts (EaE) \cite{fevry-etal-2020-entities} and Transformer Over Mention Encodings (TOME) \cite{jong2022mention} on two knowledge intensive tasks.
These models were  pretrained on Wikipedia data using entity aware losses, and their memory component focus primarily on that domain.
To make MoWE models a little more specialized on Wikipedia domain, which is known to benefit tasks such as TriviaQA, we followed \cite{roberts-etal-2020-much} and used the Salient Spam Masking (SSM) data from \cite{guu2020realm} to perform an additional number of 40K pretraining steps.

\begin{table}[h!]
\small
\centering
\begin{tabular}{lccc}
\toprule
\textbf{Model} & \textbf{TQA} & \textbf{FEVER} \\
\midrule
EaE      & 43.2 & 66.1 / 63.6 \\  
TOME 1   & 50.8 & 70.5 / 67.8\\ 
TOME 2   & \textbf{54.6} & \textbf{71.1} / 68.1 \\  
\midrule
% MoWE-Base  & & &\\
MoWE-Base + SSM  & 44.9 & 69.1 / 66.9\\
% MoWE-Large & & &\\
MoWE-Large + SSM & \textbf{50.2} & 70.5 / \textbf{68.7} \\
\bottomrule
\end{tabular}
\caption{Comparison of MoWE with EaE and TOME. Results for both models are from \cite{jong2022mention}.
Results for TQA are dev, while FEVER is dev/test. TOME 1 uses two mem. layers and TOME 2 uses two.}
\label{tab:results:mowe_vs_memaug}
\end{table}

We summarize the experimental results in Table \ref{tab:results:mowe_vs_memaug}\footnote{
For TriviaQA, we report results for the validation set only because the server used to score the test set is no longer active.}.
MoWE-Base model outperform EaE on both datasets.
MoWE-Large model outperforms both baselines on FEVER and has similar or competitive performance to TOME models on TriviaQA.

EaE and TOME models are arguably more customized solutions to these tasks. For example, EaE and TOME tackle TriviaQA as an entity linking task, where a closed set of 1M Wikipedia entities is used for ranking.
In contrast, MoWE performs open-ended answer generation, which is more flexible but also more challenging. Additionally, both EaE and TOME use specialized training procedures, including adding additional loss functions and entity or noun phrase chunking, and require k-nn tools to search relevant embeddings in their memory.
In MoWE models, the ``sparse memory'' is integrated into the model backbone and accessed seamlessly as any other model parameter. As a consequence, MoWE can be trained in a similar fashion to a T5 model with no external tools/models.

\subsection{Effectiveness of Knowledge-Driven Routing Vocabularies}
\label{sec:effect_knowledge_vocab}
In this section, we show evidence to support our conjecture that routing with large knowledge-rich vocabularies leads to better performance by varying the size of the routing vocabulary.
For the experiments in this section we use a \emph{baseline} MoWE model configuration with a fixed T51.1-Base backbone with 32K experts, yielding 15.5B sparse parameters. 
For vocabularies smaller than 1M,
we use the top-K words (by frequency in C4 dataset) from our 1M routing vocabulary described in Section \ref{sec:vocab}.
% We pretrain each model variant for 200K steps only.
We report results mainly on the TriviaQA and Natural Questions datasets and we use F1 metric instead of exact match because it is slightly less noisy and highlights the trends more clearly.

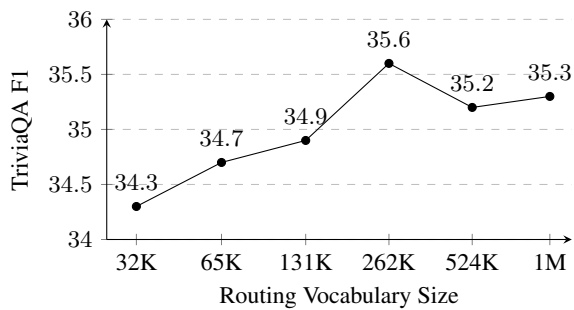
\begin{figure}[ht!]
    \centering

    \footnotesize
    \begin{tikzpicture}
    \begin{axis}[
        xlabel={Routing Vocabulary Size},
        ylabel={TriviaQA F1},
        xmin=25, xmax=1200,
        ymin=34, ymax=36.0,
        %legend pos=north west,
        xmode=log,
        xtick={32, 65, 131, 262, 524, 1000},
        xticklabels={32K, 65K, 131K, 262K, 524K, 1M},
        ymajorgrids=true,
        grid style=dashed,
        axis lines = left,
        width=\linewidth,
        height=4.5cm,
        nodes near coords,
        every node near coord/.append style={yshift=3pt}
    ]

    \addplot[color=black, scatter, mark size=1.5pt,]
        coordinates {
           (32,34.3) (65,34.7) (131,34.9) (262,35.6) (524,35.2) (1000,35.3)
        };
        
    \end{axis}
    \end{tikzpicture}    
    \caption{Performance on TriviaQA with different routing vocabulary sizes. These models are pretrained for 200K training steps.}
    \label{fig:diff_aux_voc_sizes}
\end{figure}

\begin{figure}[ht!]
    \centering

    \footnotesize
    \begin{tikzpicture}
    \begin{axis}[
        xlabel={Routing Vocabulary Size},
        ylabel={F1},
        xmin=1.3, xmax=4.7,
        ymin=32, ymax=42.0,
        %xmode=log,
        %xtick={32, 65, 1000},
        xtick={2, 3, 4},
        xticklabels={32K, 262K, 1M},
        ymajorgrids=true,
        grid style=dashed,
        axis lines = left,
        width=\linewidth,
        height=4.5cm,
        nodes near coords,
        every node near coord/.append style={yshift=3pt},
        legend style={draw=none, legend columns=-1, yshift=10pt}
    ]

    \addplot[color=black, scatter, mark size=1.5pt,]
        coordinates {
           (2,37.2) (3,39.1) (4,39.0)
        };
        
    \addplot[color=black, scatter, mark size=1.5pt, dash pattern=on 2pt off 2pt,mark=square*]
        coordinates {
           (2,32.8) (3,33.9) (4,34.3)
        };

    \legend{TriviaQA, Natural Questions}

    \end{axis}
    \end{tikzpicture}    
    \caption{Performance on TriviaQA and Natural Questions with different routing vocabulary sizes. These models are pretrained 1M training steps (longer than Figure \ref{fig:diff_aux_voc_sizes}).}
    \label{fig:tqa_nq_diff_aux_voc_sizes}
\end{figure}
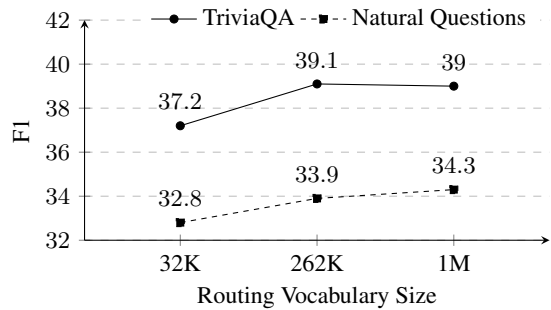

Figure \ref{fig:diff_aux_voc_sizes} shows that results progressively improve as we increase the routing vocabulary. 
These improvements are more pronounced when training for longer; see Figure \ref{fig:tqa_nq_diff_aux_voc_sizes}.
As we increase the size of the routing vocabulary, we increase the lexical-based inductive bias injected in the model via the routing function.
For TriviaQA, there is an improvement of $\sim$2 points in F1 when using routing vocabularies with size above 262K.
See Appendix \ref{appendix:ablations} for additional ablation experiments on the number of experts used.

\section{Related Work}

\textbf{Sparsely-activated Mixture-of-Experts (MoE)} models \cite{shazeer2017} increase parameter count with sublinear increases in computation cost (FLOPs) by sparsely activating modules ("experts"). Recently, Transformer-based MoE models have achieved state-of-the-art performance and efficiency wins in language \cite{lepikhin2020gshard, fedus2022switch, du2022glam, artetxe2021efficient, zoph2022stmoe}, vision \cite{riquelme2021scaling} and multimodal \cite{mustafa2022multimodal}. 

In contrast to the aforementioned MoE models, MoWE uses tens of thousands of experts; \citet{du2022glam}, for example, found diminishing performance in their MoE models beyond roughly 64 or 128 experts. To support more experts, MoWE uses a fixed routing scheme, unlike vanilla models which all rely on learned top-k routing mechanisms to assign tokens $\rightarrow$ experts, or \citet{zhou2022mixtureofexperts} who use learned top-k expert $\rightarrow$ token assignments. The MoWE routing function assigns tokens to individual experts based on their token id in an auxiliary vocabulary. This is reminiscent of Hash Layers \cite{roller2021hash}, which assigns tokens to experts based on a fixed hash bucketing, with the difference that many different token ids, based on the \emph{embedding vocabulary}, are bucketed together and assigned to individual experts. As a further consequence of the increased number of experts, we freeze the MoWE experts during finetuning to avoid both overfitting and catastrophic forgetting of knowledge acquired during pretraining.

In standard SPMD MoE implementations, experts have fixed capacity buffers and can therefore only process a fixed fraction of the input tokens, so most top-k routing models invoke an auxiliary load balancing loss \cite{shazeer2017} to encourage even distribution of tokens across experts. Because routing is fixed, MoWE expert capacity buffers can be sized according to expected token frequency. Recent work, such as \citet{gale2023megablocks} relaxes expert buffer constraints with variable expert buffer "blocks". 

MoWE models bridge the gap between MoE models and \textbf{Memory augmented models}, such as Mention Memory \cite{jong2022mention}, FILM \cite{verga-etal-2021-adaptable}, 
Entities as Experts \cite{fevry-etal-2020-entities} and Knowledge Prompts \cite{santos2022knowledge}, which call a memory bank when processing inputs. Memory models have proven effective in knowledge intensive tasks but can have few drawbacks: (1) They typically require a specialized training procedure, that differ from dense models, in order to effectively learn to use the "external" memory.
% searching the memory is a hard task, and has to be learned together (or separetely) to the model;
(2) Training data is normally very domain specific (most cases focus on Wikipedia) and, as a result, each models can only be applied to tasks that benefit from that data.

On the other hand, MoWE is simple to train -- no additional losses and no need to learn to search the memory. It seamlessly integrates with the model as there is 
no need to perform search using a nearest neighbor style tool during inference or training; the predefined routing avoids this search altogether. MoWE models can be trained on generic pretraining data (C4 in our case). The link between memory augmented and MoWE models, is that the entities are encoded into the model when identified with particular experts. However, unlike memory models, the experts/entities are small neural networks rather than embeddings.
\section{Conclusions}
We have presented MoWE, a novel neural net architecture that interpolates between the efficiency of matrix multiplication based sparsely activated MoE models and memory augmented models.
MoWE models are particularly effective at knowledge intensive tasks that require memorization and retrieval of world knowledge.
Our work brings important new findings on the use of lexical-driven routing functions in MoEs,
and hopefully invites future research on word experts.

%\pagebreak

\bibliography{anthology,custom}

\begin{thebibliography}{39}
\expandafter\ifx\csname natexlab\endcsname\relax\def\natexlab#1{#1}\fi

\bibitem[{Agarwal et~al.(2021)Agarwal, Ge, Shakeri, and
  Al-Rfou}]{agarwal-etal-2021-knowledge}
Oshin Agarwal, Heming Ge, Siamak Shakeri, and Rami Al-Rfou. 2021.
\newblock \href {https://doi.org/10.18653/v1/2021.naacl-main.278} {Knowledge
  graph based synthetic corpus generation for knowledge-enhanced language model
  pre-training}.
\newblock In \emph{Proceedings of the 2021 Conference of the North American
  Chapter of the Association for Computational Linguistics: Human Language
  Technologies}, pages 3554--3565, Online. Association for Computational
  Linguistics.

\bibitem[{Artetxe et~al.(2021)Artetxe, Bhosale, Goyal, Mihaylov, Ott, Shleifer,
  Lin, Du, Iyer, Pasunuru et~al.}]{artetxe2021efficient}
Mikel Artetxe, Shruti Bhosale, Naman Goyal, Todor Mihaylov, Myle Ott, Sam
  Shleifer, Xi~Victoria Lin, Jingfei Du, Srinivasan Iyer, Ramakanth Pasunuru,
  et~al. 2021.
\newblock Efficient large scale language modeling with mixtures of experts.
\newblock \emph{arXiv preprint arXiv:2112.10684}.

\bibitem[{Berant et~al.(2013)Berant, Chou, Frostig, and
  Liang}]{berant-etal-2013-webquestions}
Jonathan Berant, Andrew Chou, Roy Frostig, and Percy Liang. 2013.
\newblock \href {https://aclanthology.org/D13-1160} {Semantic parsing on
  {F}reebase from question-answer pairs}.
\newblock In \emph{Proceedings of the 2013 Conference on Empirical Methods in
  Natural Language Processing}, pages 1533--1544, Seattle, Washington, USA.
  Association for Computational Linguistics.

\bibitem[{Bradbury et~al.(2018)Bradbury, Frostig, Hawkins, Johnson, Leary,
  Maclaurin, Necula, Paszke, Vander{P}las, Wanderman-{M}ilne, and
  Zhang}]{jax2018github}
James Bradbury, Roy Frostig, Peter Hawkins, Matthew~James Johnson, Chris Leary,
  Dougal Maclaurin, George Necula, Adam Paszke, Jake Vander{P}las, Skye
  Wanderman-{M}ilne, and Qiao Zhang. 2018.
\newblock \href {http://github.com/google/jax} {{JAX}: composable
  transformations of {P}ython+{N}um{P}y programs}.

\bibitem[{Brown et~al.(2020)Brown, Mann, Ryder, Subbiah, Kaplan, Dhariwal,
  Neelakantan, Shyam, Sastry, Askell, Agarwal, Herbert-Voss, Krueger, Henighan,
  Child, Ramesh, Ziegler, Wu, Winter, Hesse, Chen, Sigler, Litwin, Gray, Chess,
  Clark, Berner, McCandlish, Radford, Sutskever, and
  Amodei}]{brown2020language}
Tom~B. Brown, Benjamin Mann, Nick Ryder, Melanie Subbiah, Jared Kaplan,
  Prafulla Dhariwal, Arvind Neelakantan, Pranav Shyam, Girish Sastry, Amanda
  Askell, Sandhini Agarwal, Ariel Herbert-Voss, Gretchen Krueger, Tom Henighan,
  Rewon Child, Aditya Ramesh, Daniel~M. Ziegler, Jeffrey Wu, Clemens Winter,
  Christopher Hesse, Mark Chen, Eric Sigler, Mateusz Litwin, Scott Gray,
  Benjamin Chess, Jack Clark, Christopher Berner, Sam McCandlish, Alec Radford,
  Ilya Sutskever, and Dario Amodei. 2020.
\newblock \href {http://arxiv.org/abs/2005.14165} {Language models are few-shot
  learners}.

\bibitem[{Chowdhery et~al.(2022)Chowdhery, Narang, Devlin, Bosma, Mishra,
  Roberts, Barham, Chung, Sutton, Gehrmann, Schuh, Shi, Tsvyashchenko, Maynez,
  Rao, Barnes, Tay, Shazeer, Prabhakaran, Reif, Du, Hutchinson, Pope, Bradbury,
  Austin, Isard, Gur-Ari, Yin, Duke, Levskaya, Ghemawat, Dev, Michalewski,
  Garcia, Misra, Robinson, Fedus, Zhou, Ippolito, Luan, Lim, Zoph, Spiridonov,
  Sepassi, Dohan, Agrawal, Omernick, Dai, Pillai, Pellat, Lewkowycz, Moreira,
  Child, Polozov, Lee, Zhou, Wang, Saeta, Diaz, Firat, Catasta, Wei,
  Meier-Hellstern, Eck, Dean, Petrov, and Fiedel}]{chowdhery2022palm}
Aakanksha Chowdhery, Sharan Narang, Jacob Devlin, Maarten Bosma, Gaurav Mishra,
  Adam Roberts, Paul Barham, Hyung~Won Chung, Charles Sutton, Sebastian
  Gehrmann, Parker Schuh, Kensen Shi, Sasha Tsvyashchenko, Joshua Maynez,
  Abhishek Rao, Parker Barnes, Yi~Tay, Noam Shazeer, Vinodkumar Prabhakaran,
  Emily Reif, Nan Du, Ben Hutchinson, Reiner Pope, James Bradbury, Jacob
  Austin, Michael Isard, Guy Gur-Ari, Pengcheng Yin, Toju Duke, Anselm
  Levskaya, Sanjay Ghemawat, Sunipa Dev, Henryk Michalewski, Xavier Garcia,
  Vedant Misra, Kevin Robinson, Liam Fedus, Denny Zhou, Daphne Ippolito, David
  Luan, Hyeontaek Lim, Barret Zoph, Alexander Spiridonov, Ryan Sepassi, David
  Dohan, Shivani Agrawal, Mark Omernick, Andrew~M. Dai,
  Thanumalayan~Sankaranarayana Pillai, Marie Pellat, Aitor Lewkowycz, Erica
  Moreira, Rewon Child, Oleksandr Polozov, Katherine Lee, Zongwei Zhou, Xuezhi
  Wang, Brennan Saeta, Mark Diaz, Orhan Firat, Michele Catasta, Jason Wei,
  Kathy Meier-Hellstern, Douglas Eck, Jeff Dean, Slav Petrov, and Noah Fiedel.
  2022.
\newblock \href {http://arxiv.org/abs/2204.02311} {Palm: Scaling language
  modeling with pathways}.

\bibitem[{Dai et~al.(2022)Dai, Dong, Hao, Sui, Chang, and
  Wei}]{dai-etal-2022-knowledge_neurons}
Damai Dai, Li~Dong, Yaru Hao, Zhifang Sui, Baobao Chang, and Furu Wei. 2022.
\newblock \href {https://doi.org/10.18653/v1/2022.acl-long.581} {Knowledge
  neurons in pretrained transformers}.
\newblock In \emph{Proceedings of the 60th Annual Meeting of the Association
  for Computational Linguistics (Volume 1: Long Papers)}, pages 8493--8502,
  Dublin, Ireland. Association for Computational Linguistics.

\bibitem[{de~Jong et~al.(2022)de~Jong, Zemlyanskiy, FitzGerald, Sha, and
  Cohen}]{jong2022mention}
Michiel de~Jong, Yury Zemlyanskiy, Nicholas FitzGerald, Fei Sha, and William~W.
  Cohen. 2022.
\newblock \href {https://openreview.net/forum?id=OY1A8ejQgEX} {Mention memory:
  incorporating textual knowledge into transformers through entity mention
  attention}.
\newblock In \emph{International Conference on Learning Representations}.

\bibitem[{dos Santos et~al.(2022)dos Santos, Dong, Cer, Nham, Shakeri, Ni, and
  hsuan Sung}]{santos2022knowledge}
Cicero~Nogueira dos Santos, Zhe Dong, Daniel Cer, John Nham, Siamak Shakeri,
  Jianmo Ni, and Yun hsuan Sung. 2022.
\newblock \href {http://arxiv.org/abs/2210.04726} {Knowledge prompts: Injecting
  world knowledge into language models through soft prompts}.

\bibitem[{Du et~al.(2022)Du, Huang, Dai, Tong, Lepikhin, Xu, Krikun, Zhou, Yu,
  Firat et~al.}]{du2022glam}
Nan Du, Yanping Huang, Andrew~M Dai, Simon Tong, Dmitry Lepikhin, Yuanzhong Xu,
  Maxim Krikun, Yanqi Zhou, Adams~Wei Yu, Orhan Firat, et~al. 2022.
\newblock Glam: Efficient scaling of language models with mixture-of-experts.
\newblock In \emph{International Conference on Machine Learning}, pages
  5547--5569. PMLR.

\bibitem[{Fedus et~al.(2022)Fedus, Zoph, and Shazeer}]{fedus2022switch}
William Fedus, Barret Zoph, and Noam Shazeer. 2022.
\newblock \href {http://arxiv.org/abs/2101.03961} {Switch transformers: Scaling
  to trillion parameter models with simple and efficient sparsity}.

\bibitem[{F{\'e}vry et~al.(2020)F{\'e}vry, Baldini~Soares, FitzGerald, Choi,
  and Kwiatkowski}]{fevry-etal-2020-entities}
Thibault F{\'e}vry, Livio Baldini~Soares, Nicholas FitzGerald, Eunsol Choi, and
  Tom Kwiatkowski. 2020.
\newblock \href {https://doi.org/10.18653/v1/2020.emnlp-main.400} {Entities as
  experts: Sparse memory access with entity supervision}.
\newblock In \emph{Proceedings of the 2020 Conference on Empirical Methods in
  Natural Language Processing (EMNLP)}, pages 4937--4951, Online. Association
  for Computational Linguistics.

\bibitem[{Gale et~al.(2023)Gale, Narayanan, Young, and
  Zaharia}]{gale2023megablocks}
Trevor Gale, Deepak Narayanan, Cliff Young, and Matei Zaharia. 2023.
\newblock Megablocks: Efficient sparse training with mixture-of-experts.
\newblock \emph{Proceedings of Machine Learning and Systems}, 5.

\bibitem[{Geva et~al.(2021)Geva, Schuster, Berant, and
  Levy}]{geva-etal-2021-ffn_memory}
Mor Geva, Roei Schuster, Jonathan Berant, and Omer Levy. 2021.
\newblock \href {https://doi.org/10.18653/v1/2021.emnlp-main.446} {Transformer
  feed-forward layers are key-value memories}.
\newblock In \emph{Proceedings of the 2021 Conference on Empirical Methods in
  Natural Language Processing}, pages 5484--5495, Online and Punta Cana,
  Dominican Republic. Association for Computational Linguistics.

\bibitem[{Guu et~al.(2020)Guu, Lee, Tung, Pasupat, and Chang}]{guu2020realm}
Kelvin Guu, Kenton Lee, Zora Tung, Panupong Pasupat, and Ming-Wei Chang. 2020.
\newblock {REALM}: Retrieval-augmented language model pre-training.
\newblock \emph{arXiv preprint arXiv:2002.08909}.

\bibitem[{Joshi et~al.(2017)Joshi, Choi, Weld, and
  Zettlemoyer}]{joshi-etal-2017-triviaqa}
Mandar Joshi, Eunsol Choi, Daniel Weld, and Luke Zettlemoyer. 2017.
\newblock \href {https://doi.org/10.18653/v1/P17-1147} {{T}rivia{QA}: A large
  scale distantly supervised challenge dataset for reading comprehension}.
\newblock In \emph{Proceedings of the 55th Annual Meeting of the Association
  for Computational Linguistics (Volume 1: Long Papers)}, pages 1601--1611,
  Vancouver, Canada. Association for Computational Linguistics.

\bibitem[{Kaplan et~al.(2020)Kaplan, McCandlish, Henighan, Brown, Chess, Child,
  Gray, Radford, Wu, and Amodei}]{kaplan2020scaling}
Jared Kaplan, Sam McCandlish, Tom Henighan, Tom~B. Brown, Benjamin Chess, Rewon
  Child, Scott Gray, Alec Radford, Jeffrey Wu, and Dario Amodei. 2020.
\newblock \href {http://arxiv.org/abs/2001.08361} {Scaling laws for neural
  language models}.

\bibitem[{Kudo and Richardson(2018)}]{kudo-richardson-2018-sentencepiece}
Taku Kudo and John Richardson. 2018.
\newblock \href {https://doi.org/10.18653/v1/D18-2012} {{S}entence{P}iece: A
  simple and language independent subword tokenizer and detokenizer for neural
  text processing}.
\newblock In \emph{Proceedings of the 2018 Conference on Empirical Methods in
  Natural Language Processing: System Demonstrations}, pages 66--71, Brussels,
  Belgium. Association for Computational Linguistics.

\bibitem[{Kwiatkowski et~al.(2019)Kwiatkowski, Palomaki, Redfield, Collins,
  Parikh, Alberti, Epstein, Polosukhin, Devlin, Lee, Toutanova, Jones, Kelcey,
  Chang, Dai, Uszkoreit, Le, and Petrov}]{kwiatkowski-etal-2019-natural}
Tom Kwiatkowski, Jennimaria Palomaki, Olivia Redfield, Michael Collins, Ankur
  Parikh, Chris Alberti, Danielle Epstein, Illia Polosukhin, Jacob Devlin,
  Kenton Lee, Kristina Toutanova, Llion Jones, Matthew Kelcey, Ming-Wei Chang,
  Andrew~M. Dai, Jakob Uszkoreit, Quoc Le, and Slav Petrov. 2019.
\newblock \href {https://doi.org/10.1162/tacl_a_00276} {Natural questions: A
  benchmark for question answering research}.
\newblock \emph{Transactions of the Association for Computational Linguistics},
  7:452--466.

\bibitem[{Lee et~al.(2019)Lee, Chang, and Toutanova}]{lee-etal-2019-latent}
Kenton Lee, Ming-Wei Chang, and Kristina Toutanova. 2019.
\newblock \href {https://doi.org/10.18653/v1/P19-1612} {Latent retrieval for
  weakly supervised open domain question answering}.
\newblock In \emph{Proceedings of the 57th Annual Meeting of the Association
  for Computational Linguistics}, pages 6086--6096, Florence, Italy.
  Association for Computational Linguistics.

\bibitem[{Lepikhin et~al.(2020)Lepikhin, Lee, Xu, Chen, Firat, Huang, Krikun,
  Shazeer, and Chen}]{lepikhin2020gshard}
Dmitry Lepikhin, HyoukJoong Lee, Yuanzhong Xu, Dehao Chen, Orhan Firat, Yanping
  Huang, Maxim Krikun, Noam Shazeer, and Zhifeng Chen. 2020.
\newblock Gshard: Scaling giant models with conditional computation and
  automatic sharding.
\newblock \emph{arXiv preprint arXiv:2006.16668}.

\bibitem[{Meng et~al.(2022)Meng, Bau, Andonian, and
  Belinkov}]{meng2022locating}
Kevin Meng, David Bau, Alex~J Andonian, and Yonatan Belinkov. 2022.
\newblock \href {https://openreview.net/forum?id=-h6WAS6eE4} {Locating and
  editing factual associations in {GPT}}.
\newblock In \emph{Advances in Neural Information Processing Systems}.

\bibitem[{Mustafa et~al.(2022)Mustafa, Riquelme, Puigcerver, Jenatton, and
  Houlsby}]{mustafa2022multimodal}
Basil Mustafa, Carlos Riquelme, Joan Puigcerver, Rodolphe Jenatton, and Neil
  Houlsby. 2022.
\newblock Multimodal contrastive learning with limoe: the language-image
  mixture of experts.
\newblock \emph{arXiv preprint arXiv:2206.02770}.

\bibitem[{Raffel et~al.(2020)Raffel, Shazeer, Roberts, Lee, Narang, Matena,
  Zhou, Li, and Liu}]{2020t5}
Colin Raffel, Noam Shazeer, Adam Roberts, Katherine Lee, Sharan Narang, Michael
  Matena, Yanqi Zhou, Wei Li, and Peter~J. Liu. 2020.
\newblock \href {http://jmlr.org/papers/v21/20-074.html} {Exploring the limits
  of transfer learning with a unified text-to-text transformer}.
\newblock \emph{Journal of Machine Learning Research}, 21(140):1--67.

\bibitem[{Riquelme et~al.(2021)Riquelme, Puigcerver, Mustafa, Neumann,
  Jenatton, Susano~Pinto, Keysers, and Houlsby}]{riquelme2021scaling}
Carlos Riquelme, Joan Puigcerver, Basil Mustafa, Maxim Neumann, Rodolphe
  Jenatton, Andr{\'e} Susano~Pinto, Daniel Keysers, and Neil Houlsby. 2021.
\newblock Scaling vision with sparse mixture of experts.
\newblock \emph{Advances in Neural Information Processing Systems},
  34:8583--8595.

\bibitem[{Roberts et~al.(2022)Roberts, Chung, Levskaya, Mishra, Bradbury,
  Andor, Narang, Lester, Gaffney, Mohiuddin et~al.}]{roberts2022scaling}
Adam Roberts, Hyung~Won Chung, Anselm Levskaya, Gaurav Mishra, James Bradbury,
  Daniel Andor, Sharan Narang, Brian Lester, Colin Gaffney, Afroz Mohiuddin,
  et~al. 2022.
\newblock Scaling up models and data with t5x and seqio.
\newblock \emph{arXiv preprint arXiv:2203.17189}, 13.

\bibitem[{Roberts et~al.(2020)Roberts, Raffel, and
  Shazeer}]{roberts-etal-2020-much}
Adam Roberts, Colin Raffel, and Noam Shazeer. 2020.
\newblock \href {https://doi.org/10.18653/v1/2020.emnlp-main.437} {How much
  knowledge can you pack into the parameters of a language model?}
\newblock In \emph{Proceedings of the 2020 Conference on Empirical Methods in
  Natural Language Processing (EMNLP)}, pages 5418--5426, Online. Association
  for Computational Linguistics.

\bibitem[{Roller et~al.(2021)Roller, Sukhbaatar, Szlam, and
  Weston}]{roller2021hash}
Stephen Roller, Sainbayar Sukhbaatar, Arthur Szlam, and Jason~E Weston. 2021.
\newblock \href {https://openreview.net/forum?id=lMgDDWb1ULW} {Hash layers for
  large sparse models}.
\newblock In \emph{Advances in Neural Information Processing Systems}.

\bibitem[{Shazeer et~al.(2017)Shazeer, Mirhoseini, Maziarz, Davis, Le, Hinton,
  and Dean}]{shazeer2017}
Noam Shazeer, *Azalia Mirhoseini, *Krzysztof Maziarz, Andy Davis, Quoc Le,
  Geoffrey Hinton, and Jeff Dean. 2017.
\newblock \href {https://openreview.net/forum?id=B1ckMDqlg} {Outrageously large
  neural networks: The sparsely-gated mixture-of-experts layer}.
\newblock In \emph{International Conference on Learning Representations}.

\bibitem[{Thorne et~al.(2018)Thorne, Vlachos, Christodoulopoulos, and
  Mittal}]{thorne-etal-2018-fever}
James Thorne, Andreas Vlachos, Christos Christodoulopoulos, and Arpit Mittal.
  2018.
\newblock \href {https://doi.org/10.18653/v1/N18-1074} {{FEVER}: a large-scale
  dataset for fact extraction and {VER}ification}.
\newblock In \emph{Proceedings of the 2018 Conference of the North {A}merican
  Chapter of the Association for Computational Linguistics: Human Language
  Technologies, Volume 1 (Long Papers)}, pages 809--819, New Orleans,
  Louisiana. Association for Computational Linguistics.

\bibitem[{Touvron et~al.(2023)Touvron, Lavril, Izacard, Martinet, Lachaux,
  Lacroix, Rozière, Goyal, Hambro, Azhar, Rodriguez, Joulin, Grave, and
  Lample}]{touvron2023llama}
Hugo Touvron, Thibaut Lavril, Gautier Izacard, Xavier Martinet, Marie-Anne
  Lachaux, Timothée Lacroix, Baptiste Rozière, Naman Goyal, Eric Hambro,
  Faisal Azhar, Aurelien Rodriguez, Armand Joulin, Edouard Grave, and Guillaume
  Lample. 2023.
\newblock \href {http://arxiv.org/abs/2302.13971} {Llama: Open and efficient
  foundation language models}.

\bibitem[{Vaswani et~al.(2017)Vaswani, Shazeer, Parmar, Uszkoreit, Jones,
  Gomez, Kaiser, and Polosukhin}]{transformer}
Ashish Vaswani, Noam Shazeer, Niki Parmar, Jakob Uszkoreit, Llion Jones,
  Aidan~N. Gomez, \L{}ukasz Kaiser, and Illia Polosukhin. 2017.
\newblock Attention is all you need.
\newblock In \emph{Proceedings of the 31st International Conference on Neural
  Information Processing Systems}, NIPS'17, page 6000–6010, Red Hook, NY,
  USA. Curran Associates Inc.

\bibitem[{Verga et~al.(2021)Verga, Sun, Baldini~Soares, and
  Cohen}]{verga-etal-2021-adaptable}
Pat Verga, Haitian Sun, Livio Baldini~Soares, and William Cohen. 2021.
\newblock \href {https://doi.org/10.18653/v1/2021.naacl-main.288} {Adaptable
  and interpretable neural {M}emory{O}ver symbolic knowledge}.
\newblock In \emph{Proceedings of the 2021 Conference of the North American
  Chapter of the Association for Computational Linguistics: Human Language
  Technologies}, pages 3678--3691, Online. Association for Computational
  Linguistics.

\bibitem[{Wang et~al.(2019{\natexlab{a}})Wang, Pruksachatkun, Nangia, Singh,
  Michael, Hill, Levy, and Bowman}]{wang2019superglue}
Alex Wang, Yada Pruksachatkun, Nikita Nangia, Amanpreet Singh, Julian Michael,
  Felix Hill, Omer Levy, and Samuel Bowman. 2019{\natexlab{a}}.
\newblock Superglue: A stickier benchmark for general-purpose language
  understanding systems.
\newblock \emph{Advances in neural information processing systems}, 32.

\bibitem[{Wang et~al.(2019{\natexlab{b}})Wang, Pruksachatkun, Nangia, Singh,
  Michael, Hill, Levy, and Bowman}]{superglue}
Alex Wang, Yada Pruksachatkun, Nikita Nangia, Amanpreet Singh, Julian Michael,
  Felix Hill, Omer Levy, and Samuel~R. Bowman. 2019{\natexlab{b}}.
\newblock \emph{SuperGLUE: A Stickier Benchmark for General-Purpose Language
  Understanding Systems}. Curran Associates Inc., Red Hook, NY, USA.

\bibitem[{Xue et~al.(2022)Xue, Barua, Constant, Al-Rfou, Narang, Kale, Roberts,
  and Raffel}]{xue2022byt5}
Linting Xue, Aditya Barua, Noah Constant, Rami Al-Rfou, Sharan Narang, Mihir
  Kale, Adam Roberts, and Colin Raffel. 2022.
\newblock \href {http://arxiv.org/abs/2105.13626} {Byt5: Towards a token-free
  future with pre-trained byte-to-byte models}.

\bibitem[{Zhang et~al.(2022)Zhang, Lin, Liu, Li, Sun, and
  Zhou}]{zhang-etal-2022-moefication}
Zhengyan Zhang, Yankai Lin, Zhiyuan Liu, Peng Li, Maosong Sun, and Jie Zhou.
  2022.
\newblock \href {https://doi.org/10.18653/v1/2022.findings-acl.71}
  {{M}o{E}fication: Transformer feed-forward layers are mixtures of experts}.
\newblock In \emph{Findings of the Association for Computational Linguistics:
  ACL 2022}, pages 877--890, Dublin, Ireland. Association for Computational
  Linguistics.

\bibitem[{Zhou et~al.(2022)Zhou, Lei, Liu, Du, Huang, Zhao, Dai, Chen, Le, and
  Laudon}]{zhou2022mixtureofexperts}
Yanqi Zhou, Tao Lei, Hanxiao Liu, Nan Du, Yanping Huang, Vincent Zhao, Andrew
  Dai, Zhifeng Chen, Quoc Le, and James Laudon. 2022.
\newblock \href {http://arxiv.org/abs/2202.09368} {Mixture-of-experts with
  expert choice routing}.

\bibitem[{Zoph et~al.(2022)Zoph, Bello, Kumar, Du, Huang, Dean, Shazeer, and
  Fedus}]{zoph2022stmoe}
Barret Zoph, Irwan Bello, Sameer Kumar, Nan Du, Yanping Huang, Jeff Dean, Noam
  Shazeer, and William Fedus. 2022.
\newblock \href {http://arxiv.org/abs/2202.08906} {St-moe: Designing stable and
  transferable sparse expert models}.

\end{thebibliography}
\bibliographystyle{acl_natbib}

\appendix
\section{MoWE Setup}
\label{sec:appendix:mowe_setup}

Or main experiments on MoWE-Base and MoWE-Large use an architecture with four MoWE-layers in total.
Two in the encoder and two in the decoder and parameters are shared across all MoWE layers.
Those layers are placed at Transformer blocks 5 and 10 in the Base model,
and at blocks 9 and 17 in the large model.
We placed MoWE layers towards the middle of the encoder (decoder) because: 
(1) they receive a representation of the token that is already somewhat contextualized;
(2) after the MoWE layer, there are still multiple Transformer Blocks that can benefit from the output of that layer.
Unless otherwise informed,
each MoWE layer contains 32K experts and the routing vocabulary has \textasciitilde1M entries.

Our current implementation of MoWE was coded in Jax \cite{jax2018github} on top of the T5X \cite{roberts2022scaling} framework. \footnote{\url{https://github.com/google-research/t5x}}

Some additional configurations are provided in the following sections.

\subsection{Configuration of Frequency Buckets, Expert Blocks and Experts}
\label{appendix:freq_bucket}
We split the vocabulary into four different frequency buckets. Token frequency was computed using a sample from our pretraining dataset.
The MoWE layer does not process the top 16 most frequent tokens in the routing vocabulary, i.e. those tokens ids are never routed to an expert.
These tokens are punctuation marks and other non-content words and we estimate they can represent up to 28\% of the tokens in a batch.
This speeds up the training time and does not hurt downstream performance, as these tokens are not content words.
The configuration of the four frequency buckets is described in Table \ref{tab:results:freq_buckets}.
Using this configuration,
we get a model with \textasciitilde31B parameters in the case of the Base model and \textasciitilde45.5B sparse parameters in the case of the Large model. 
The difference in the number of parameters is due to the use of different MLP projection dimensions (see Table \ref{tab:results:freq_buckets}) and the token embedding size,
which is 768 in Base and 1024 in Large.

\begin{table*}[h!]
\small
\centering
\begin{tabular}{lcccc}
\toprule
Configuration & \textbf{Bucket 1} & \textbf{Bucket 2} & \textbf{Bucket 3} & \textbf{Bucket 4} \\
\midrule
%Avrg. \% of tokens in batch & 23.5 & 18.42 & 7.0 & 22.8 \\
Num. of routing ids covered  & 128 & 880 &  1024 & $\sim2^{20}$ \\
Num. of Expert Blocks      & 128 & 128 & 128 & 128 \\
Num. of Experts per Block & 1   & 7   & 8 & 235 \\
Total Num. of Experts     & 128 & 896 & 1024 & 30080\\
Expert MLP dimension (Base)      & 2048 & 2048 & 1024 & 512 \\
Expert MLP dimension (Large)      & 2816 & 2816 & 1536 & 512 \\
\bottomrule
\end{tabular}
\caption{Configuration of experts and blocks in the four frequency buckets used in MoWE-Base (31B) and MoWE-Large (45B). A routing vocabulary of 1M token ids is considered.}
\label{tab:results:freq_buckets}
\end{table*}

\begin{table*}[h!]
\small
\centering
\begin{tabular}{lcccc}
\toprule
Configuration & \textbf{Bucket 1} & \textbf{Bucket 2} & \textbf{Bucket 3} & \textbf{Bucket 4} \\
\midrule
%Avrg. \% of tokens in batch & 23.5 & 18.42 & 7.0 & 22.8 \\
Num. of routing ids covered     & 128 & 880 &  1024 & $\sim2^{20}$ \\
Num. of Expert Blocks           & 64 & 64 & 64 & 64 \\
Num. of Experts per Block       & 1  & 1  & 1 & 128 \\
Total Num. of Experts           & 64 & 64 & 64 & 8192\\
Expert MLP dimension  & 2048 & 2048 & 2048 & 96 \\
\bottomrule
\end{tabular}
\caption{Configuration of experts and blocks in the four frequency buckets used in the MoWE-Base model with 2B parameters. A routing vocabulary of 1M token ids is considered.}
\label{tab:results:freq_buckets_mowebase2b}
\end{table*}

Notice in Table \ref{tab:results:freq_buckets} that for buckets 1 to 3 we use one expert per token.
In this configuration,
in bucket 4 the experts are shared for multiple tokens.
This bucket contains mainly low frequency tokens,
which are the majority in the vocabulary.
Additionally,
due to the large number of experts in this bucket, 
the Expert Blocks are implemented as lookup tables.
Although we believe the current configuration is not optimal and can be improved,
it already produces efficient models.

In Table \ref{tab:results:freq_buckets_mowebase2b},
we detail the configuration of the four frequency buckets and respective expert number and sizes for the MoWE-Base model with 2B parameters which we refer in Section \ref{sec:comparison_to_moe}.

\subsection{Additional hyperparameters}
For pretraining MoWE models, we used the default T5x hyperparamters for T5.1.1.
Unless otherwise mentioned, pretraining is performed for roughly 1 trillion tokens -- 1M steps, with batch size 2048 and input sequence length of 512 tokens; the target sequence length is 114.

For downstream task we normally use batch sizes of 256 or 512.
For most datasets, a learning rate of 1e-4 and dropout rate of 0.05 gave the best results.
The main exception is SuperGLUE and Fever datasets, which work better with LRs between 1e-3 and 5e-4.

\section{Additional ablation experiments}
\label{appendix:ablations}
In this section we present additional ablation experiments on different architectural choices of MoWE.
In all experiments, we pretrain the models for 200K steps.

\subsubsection{Effect of Number of Experts}
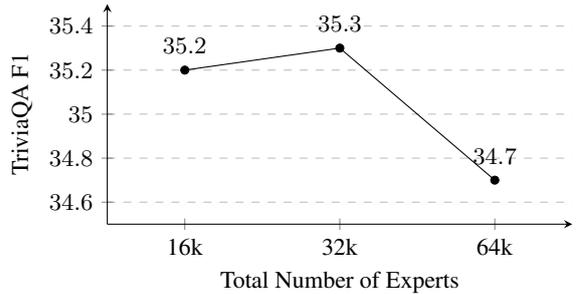
\begin{figure}[]
    \centering
    \footnotesize

    \begin{tikzpicture}
    \begin{axis}[
        xlabel={Total Number of Experts},
        ylabel={TriviaQA F1},
        %xmode=log,
        xmin=1.5, xmax=4.5,
        ymin=34.5, ymax=35.5,
        %legend pos=north west,
        xtick={2, 3, 4},
        xticklabels={16k, 32k, 64k},
        ymajorgrids=true,
        grid style=dashed,
        axis lines = left,
        width=\linewidth,
        height=4.5cm,
        nodes near coords,
        every node near coord/.append style={yshift=3pt}
    ]

    \addplot[color=black, scatter, mark size=1.5pt,]
        coordinates {
           (2,35.2) (3,35.3) (4,34.7)
        };
        
    \end{axis}
    \end{tikzpicture}
    
    \caption{Performance on TriviaQA of different MoWE-baseline models where we fix the routing vocabulary to 1M size and vary the number of experts.}
    \label{fig:fix_aux_voc_sizes_and_diff_num_experts}
\end{figure}

We present two additional experiments on how the number of experts affect MoWE performance.
First, we check the impact of varying the number of experts between 16K, 32K and 64K while keeping fixed the routing vocabulary to 1M size and the model size to 15.5B.
In Fig. \ref{fig:fix_aux_voc_sizes_and_diff_num_experts} we see that 32K experts seems to be a sweet spot in terms of number of experts for MoWE.
Using a larger number of smaller experts is preferable because is is more memory efficient and also speeds up our lookup table implementation of Expert Blocks in frequency bucket 4.

\begin{figure}[ht!]
    \centering

    \footnotesize
    \begin{tikzpicture}
    \begin{axis}[
        xlabel={Number of Experts == Routing Vocab. Size},
        ylabel={TriviaQA F1},
        xmin=25, xmax=1200,
        ymin=32.2, ymax=34.8,
        %legend pos=north west,
        xmode=log,
        xtick={32, 65, 131, 262, 524, 1048},
        xticklabels={32K, 65K, 131K, 262K, 524K, 1M},
        ymajorgrids=true,
        grid style=dashed,
        axis lines = left,
        width=\linewidth,
        height=4.5cm,
        nodes near coords,
        every node near coord/.append style={yshift=3pt}
    ]

    \addplot[color=black, scatter, mark size=1.5pt,]
        coordinates {
           (32,34.5) (65,34.2) (131,34.1) (262,34.0) (524,33.7) (1048,32.6)
        };
        
    \end{axis}
    \end{tikzpicture}    
    \caption{Performance on TriviaQA of different MoWE-baseline models where the number of experts match the routing vocabulary.}
    \label{fig:diff_aux_voc_sizes_and_num_experts}
\end{figure}
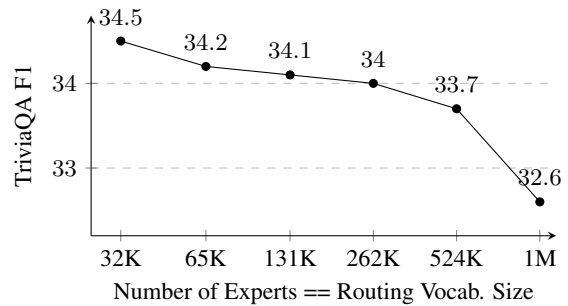

Next, we evaluate MoWE performance when we increase the number of experts to match the size of large routing vocabularies.
We keep the total number of sparse parameters fixed by decreasing the size of the experts in each experiment.
Therefore, when using 1M experts, the MLP dim of each expert is 8, while the MLP proj dimension when using 32K experts is 256.
To the best of our knowledge, this is the first time that a Transformer-based MoE model is trained with up to a million experts, demonstrating that our proposed solutions to implement MoWE is quite robust.

In Fig. \ref{fig:diff_aux_voc_sizes_and_num_experts} we show results for increasing MoWE-baseline for up to 1M experts.
We see a progressive degradation in performance when matching the number of experts to the size of the vocabulary.
We believe this is mainly due to two factors: (1) the number of training updates that each experse receive becomes increasingly sparse; (2) the size of the experts are decreased.

\subsubsection{Impact of the number of MoWE Layers}
In Table \ref{tab:results:num_mowe_layers},
we show the impact of using a different number of MoWE-Layers in encoder and decoder.
All models were trained for 200K steps.
We can see in Table \ref{tab:results:num_mowe_layers} that going from one to two layers in the encoder gives a significant gain in EM (31.0 -> 31.6). However, going from 2 to 3 layers does not give improvements on EM.
Adding MoWE layers to the decoder improves the performance, specially when using 2 layers in the encoder.

\begin{table}[h!]
\small
\centering
\begin{tabular}{cccc}
\toprule
\multicolumn{2}{c}{\textbf{\# MoWE Layers}} & \textbf{EM} & \textbf{F1} \\
\textbf{Encoder} & \textbf{Decoder} & & \\
\midrule
1 & 0 & 31.0 & 36.3 \\
2 & 0 & 31.6 & 36.9 \\
3 & 0 & 31.5 & 37.1 \\
1 & 1 & 31.4 & 36.7 \\
2 & 1 & 32.4 & 37.5 \\
2 & 2 & 33.1 & 38.4 \\
\bottomrule
\end{tabular}
\caption{Impact of the number of MoWE layers on TriviaQA for Base model. Expert parameters are shared across MoWE layers in the encoder.}
\label{tab:results:num_mowe_layers}
\end{table}

In Table \ref{tab:results:expert_size} we show results on using different expert sizes in each of the four bucket sizes.
A single MoWE layer is used, and it is located in the encoder.
We start with a configuration where the experts in Bucket 1 has experts with MLP dimension 512, and sequentially half the value for the next consecutive bucket.
This results in a model with 3.9B sparse params, whose performance on TriviaQA is presented in the firs row of Table \ref{tab:results:expert_size}.
The the following rows, we consecutively double the size of the expert in each bucket, which doubles the total number of sparse parameters.
There is a consistent improvement of 1 point in EM when doubling the model size.
We believe the increase would be larger if we pretrained the model for 1M steps instead of 200K steps.

\begin{table}[h!]
\small
\centering
\begin{tabular}{rcccccc}
\toprule
\textbf{\# sparse } & \multicolumn{4}{c}{\textbf{MLP Dim. of Experts}} & \textbf{EM} & \textbf{F1} \\
\textbf{params} & \multicolumn{4}{c}{\textbf{in each Frequency Bucket}} & & \\
& B1 & B2 & B3 & B4  \\
\midrule
3.9B  & 512 & 256 & 128 & 64      & 28.5 & 33.7 \\
7.8B  & 1024 & 512 & 256 & 128   & 29.6 & 34.8 \\
15.5B & 2048 & 1024 & 512 & 256  & 30.0 & 35.3 \\
31.0B  & 2048 & 2048 & 1024 & 512 & 31.0 & 36.3 \\
\bottomrule
\end{tabular}
\caption{Impact on TriviaQA EM and F1 of using different expert sizes in the four different buckets.}
\label{tab:results:expert_size}
\end{table}

\subsubsection{Freezing vs Unfreezing Experts During Finetuning}
\label{sec:appendix:freezing}
MoWE-Base on TriviaQA gets EM of 37.7 when freezing the experts during finetuning.
When we allow the update of experts during finetuning, EM drops by 5 points to 33.5.

\section{Metrics and Baseline Setup}
\label{sec:appendix:metric_baselines}

We use the following metrics in our experiments:
for TriviaQA, WebQuestions and Natural Question we mostly report results in terms of Exact Match (EM), except for some ablation experiments, where we report results in terms of F1.
For Fever dataset, we report the accuracy in both validation and test sets.
For SuperGLUE, following previous works \cite{2020t5,xue2022byt5},
we finetune MoWE models on a mixture of all tasks in the benchmark,
select the best result per task and present the average validation set scores over all tasks.

We use the MoE-Top2 implementation from T5x framework in our comparative experiments.
Dense and sparse layers are interleaved, which results in a total of 12 sparse layers:
6 in the encoder and 6 in the decoder.
We use Top-2 routing and most hyperparameters are default, except for expert dropout (0.3) and learning rate during finetuning,
which we set to 5e-4 for QA tasks and Fever.
For SuperGLUE, we follow the recomendation from ST-MoE paper and used a larger learning rate (1e-3) and small batch size (256), except for the model with 512 experts, for which we used batch size of 512. 
https://github.com/google-research/t5x/

\section{Example of Entries from Knowledge Rich Vocabulary}
\label{sec:appendix:knowledge_vocab}

Top 50 word, by frequency in C4, in the routing vocabulary:
\textit{'isn',
 'aren',
 '…',
 '3d',
 '1st',
 'whilst',
 'copyright',
 'creates',
 '2nd',
 'tells',
 'adds',
 'wet',
 '3rd',
 '·',
 'likes',
 'filling',
 'yours',
 '\^',
 'accordance',
 '4th',
 'amongst',
 'sees',
 '20th',
 'mp3',
 '5th',
 'woods',
 '19th',
 'tx',
 'toy',
 'solely',
 'thinks',
 '21st',
 'sits',
 'asks',
 '10th',
 'receives',
 'worlds',
 '6th',
 'singles',
 'blues',
 'tops',
 'inn',
 'lean',
 'mills',
 '7th',
 'ranges',
 'bears',
 'newer',
 '8th',
 'node'.
} 

In the top 50 words by frequency, we still see many words that are variations of common words, like \textit{"sees"}. However, the quality of the vocabulary improves significantly later in the rank. For instance,
this are the top 50 after position 6000 of 1M:
\textit{'consignment',
 'billboards',
 'primal',
 'discrepancy',
 'callback',
 'freeware',
 'horticulture',
 'jb',
 's8',
 'aspirants',
 'commemorative',
 'brisk',
 'arched',
 'pondering',
 'fluff',
 'diwali',
 'landline',
 'wilder',
 'apocalyptic',
 'patchwork',
 'airs',
 'stagnant',
 '412',
 'watery',
 'hospitalization',
 'mccoy',
 'serbian',
 'paprika',
 'headsets',
 'deserts',
 'pulley',
 'orthopaedic',
 'disparity',
 'egyptians',
 'painfully',
 'kenyan',
 'bale',
 'condemnation',
 'deportation',
 'incline',
 'perfumes',
 'undergraduates',
 'favoured',
 'pvp',
 'bbb',
 'lyons',
 'fremont',
 'eurozone',
 'afl',
 'monogram'.
 }
 
 More work can definitely be done to improve the routing vocabulary, but we wanted to keep it simple for our experiments.

\end{document}